\RequirePackage{fix-cm}
\documentclass[smallextended]{svjour3}       
\smartqed  
\usepackage{graphicx}
\usepackage{natbib}
\usepackage[ruled,vlined]{algorithm2e}
\usepackage{hyperref}
\usepackage{amssymb}
\usepackage{amsmath}

\usepackage{caption}
\usepackage{subcaption}

\newcommand{\Loss}{\mathcal{L}}

\begin{document}

\title{Tax Evasion Risk Management Using a Hybrid Unsupervised Outlier Detection Method} 
\thanks{This research is supported by Ministry of Education, Science and Technological Development and Tax Administration of Republic of Serbia}

\titlerunning{The HUNOD Method for Tax Evasion Risk Management}        

\author{
        Milo\v{s} Savi\'c\and
        Jasna Atanasijevi\'c\and
        Du\v{s}an Jakoveti\'c \and Nata\v{s}a Kreji\'c
}


\institute{ \at
              University of Novi Sad, Faculty of Sciences, Department of Mathematics and Informatics \\ Trg Dositeja Obradovi\'ca 4, 21000 Novi Sad, Serbia\\
              \email{\{svc, jasna.atanasijevic, dusan.jakovetic, natasa\}@dmi.uns.ac.rs} The authors acknowledge financial support of the Ministry of Education, Science and Technological Development of the Republic of Serbia (Grant No. 451-03-9/2021-14/ 200125)            
}

\date{Received: date / Accepted: date}

\maketitle

\begin{abstract}
Big data methods are becoming an important tool for tax fraud detection around the world. 
Unsupervised learning approach is the dominant framework due to the lack of 
label and ground truth in corresponding data sets, 
although these methods suffer from lower interpretability 
and precision compared with supervised 
approaches. In contrast to prior research works, we examine the possibility of a hybrid unsupervised method 
for tax evasion risk management that is able to internally validate 
and explain outliers detected in a given tax dataset. 
The proposed method, HUNOD (Hybrid UNsupervised Outlier Detection)~\footnote{HUNOD is an abbreviation for ``Hybrid UNsupervised Outlier Dection''}, 
combines clustering and representation learning for robust outlier
detection, additionally allowing its users
to incorporate relevant domain knowledge into both 
constituent outlier detection approaches in order to identify outliers
relevant for a given economic context.
The interpretability of obtained outliers is achieved by 
training explainable-by-design surrogate models over internally validated outliers.
The experimental evaluation of the HUNOD method is conducted on 
two datasets derived from the database on individual personal income tax 
declarations collected by the Tax Administration of Serbia. The obtained
results show that the method indicates between 90\% and 
98\% internally validated outliers depending on the clustering configuration
and employed regularization mechanisms for representational learning.

\end{abstract}

\raggedbottom

\section{Introduction}
\label{intro}
Tax evasion and tax avoidance represent a big challenge for authorities in all countries of the world. The loss from tax irregularities is estimated at 3.2\% of GDP in OECD countries~\citep{Buehn2016}.  Tax evasion reduces tax base and related public resources for provision of public goods and erodes fiscal equity. It is empirically proven that grey economy including tax evasion is higher in the countries with a lower level of per capita income~\citep{Schneider2005}. The same literature shows that the level of tax evasion is higher in the countries where the tax system structure implies more reliance on the taxation of production factors, especially on personal income tax, than the taxation of consumption as it is easier to avoid the first type of tax. 

The traditional way for improving tax compliance by tax audit is costly and limited in terms of outreach given the huge population of taxpayers and small and expensive capacity of tax controllers in tax administration. The domain knowledge, intuition and experience of tax auditors used for selection of tax payers for audit takes no advantage of rich information set registered in tax administration databases. 
It is also observed that the digitalization of government services is essential in tax avoidance decrease~\citep{UYAR2021}.
The recent evolution in data science found its vast applications in tax administrations around the world in support to traditional methods for detecting fraudulent behavior and better use of resources dedicated to tax collection. Large amounts of available tax data also enable development of mathematical
models to study and understand tax fraud dynamics~\citep{Chica2021}. 
Machine learning (ML) algorithms on administrative big data sets are increasingly being developed to improve risk management by tax authorities and help prevention and detection of tax evasion cases. 
However, several challenges remain that have not been completely addressed, as detailed below.


\subsection{Motivation and Research Problems}

The detection of tax evasion can be effectively approached by methods based on supervised ML techniques. However, those methods require large training datasets containing data instances corresponding to both verified tax evasion cases and regular (compliant) tax behaviour. As indicated in many previous studies reviewed in the related work section, such datasets are usually not available, especially in developing countries. The second problem with supervised ML approaches could be a small number of frauds identified by administrative tax controls that are recorded in the training dataset. If recorded cases are not representative of the entire population then the trained supervised model will be biased, i.e., it will have a high precision, but a low recall.

Due to the lack of availability of labelled data and ground truth that would enable supervised learning, the detection of tax evasion is usually approached by unsupervised ML methods based on anomaly detection algorithms. Compared with supervised approaches, unsupervised methods are less precise: they will not identify only tax evasion cases, but also indicate business entities with irregular and suspicious tax behaviour including also dishonest tax payers. Therefore, unsupervised methods are suitable as a decision support tool in tax evasion risk management systems enabling better prioritization of tax controls and improving efficiency of tax collection. Second, a tax evasion risk management system based on accurate unsupervised learning may 
lead to a more efficient use of resources when operationally treating identified risks~\citep{Sebtaoui2020}.

The focus of this paper is on unsupervised ML methods for identifying suspicious tax behaviour at the level of individual business entities. Such entities can be considered as point anomalies in tax databases, so outlier detection algorithms are the most natural choice in this case. One of the main issues with outlier detection algorithms (and unsupervised machine learning in general) is the validation of obtained results~\citep{Goix16}. As our primary research problem, we examine the possibility of a hybrid outlier detection method that is able to internally validate identified outliers by combining outlier detection approaches based on radically different machine learning designs.

Most existing anomaly detection studies focus on devising detection models only, ignoring the capability of providing explanation of the identified anomalies~\citep{Pang2020}. On the other hand, explainability is of high importance in this context, as it could be used to improve the tax collection and perhaps more importantly, provide guidance for systematic changes in taxation laws.  
Highly explainable features supported by the domain knowledge can be exploited to target a specific type of risky tax payers in aim to prevent dishonest fraudulent behaviour in declaring taxes. Outlier detection models with explanatory capabilities may increase both ease-of-use and usefulness of task evasion risk management systems~\citep{Wang2019}. Thus, we investigate how to devise a tax-related outlier detection approach having strong explanatory capabilities.

Finally, it is highly desirable that ML-based fraud detection models offer a systematic way of incorporating and harnessing domain knowledge in the model, rather than adopting a black-box approach, in order to make detection models more robust and precise. Such domain knowledge incorporation should be user friendly in the sense that domain experts need to invest a low-to-moderate-effort manual work to encode prior domain knowledge in the models. Thus, the last research problem examined in this paper is how to devise a hybrid tax-related outlier detection method that can be easily enhanced with relevant domain knowledge.

\subsection{Methodological Contributions}

In this paper, we propose a hybrid outlier detection method based on unsupervised learning that can incorporate relevant tax-related domain knowledge. The method is abbreviated as HUNOD (Hybrid UNsupervised Outlier Detection) method from now on. The HUNOD method was developed starting from a tax dataset of individual business entities that was derived from tax declarations for all types of personal income.

The HUNOD method independently applies two unsupervised anomaly detection methods: one based on $K$-means to detect outliers from the clustering structure of the data and another based on an autoencoder detecting anomalies according to learned latent representations of non-suspicious tax behaviour. Outliers found by the these two methods are then cross-checked in order to provide a final set of the internally-validated anomalies. 

HUNOD allows for a systematic and user-friendly way of incorporating domain knowledge in the outlier detection process. A user incorporates domain knowledge by defining thresholds on relevant scoring indicators prior to learning latent representations and by weighting various data features 
prior to identifying clustering structures. Those numeric quantities backed by economic theory reflect common business practices related to tax legal framework (e.g. tax arbitrage across different income categories with different tax burden).

An explainable-by-design decision tree model is trained over a surrogate dataset in which data instance are labelled such that internally-validated outliers are separated from other business entities. This last step of HUNOD allows domain experts to externally validate (manually or through other means) identified anomalies. Thus, the main advantages of the proposed method are the following: domain knowledge is incorporated at the very beginning in the method to increase its accuracy, the hybrid nature of HUNOD provides internal validation of identified outliers, while its explainability capabilities support external verification of the outliers by tax experts.

\subsection{Case Study}

Empirical evaluation of HUNOD is conducted in cooperation with the tax authority of Serbia which has provided a depersonalised dataset of individual tax declarations for the purpose of this research. Serbia is a middle income country with GDP per capita in 2019 of 18,179 international dollars based on purchasing power parity  (41\% of EU-27 average) and with a relatively high level of tax avoidance. 
The overall level of the shadow economy (unregistered business transactions) is estimated at about 15.4\% of GDP in 2017 using a perception based survey conducted in 2017~\citep{NALED}. The dominant form of shadow economy activity consists of undeclared labor costs, implying the entire or partial payment of salaries in cash. Tax system in Serbia is assessed as very complex and unfair. There is a large number of taxes, personal income from various sources is taxed differently. So, there are many tax breaks available~\citep{Schneider2015}. Personal income taxes and contributions represent the major single source of consolidated public revenue  (38\% in 2020) following by the VAT (25\% in 2020). Technically, most declaration and payment of both income tax and mandatory social security contributions is submitted and paid by payer of income (almost 99.9\% of all declarations). 

\subsection{Paper Organization}

The remainder of the paper is organized as follows. In Section 2  we present the related work in the area of tax fraud detection and give a short overview of methods dealing with challenges related to the anomaly detection on unlabeled data like low explanation power. We  also specify in more details the novelty of the approach presented in this paper in relation to the existing works. The HUNOD method is presented in Section 3 in details. Section 4 contains a description of the dataset used for empirical testing of the method. The obtained results are presented and discussed in Section 5. Section 6 concludes the paper and outlines possible directions for further research.

\section{Related Works and Our Contributions}
\label{rw}

According to the type of ML technique employed, existing ML-based methods to identify tax evasion, tax avoidance or 
more broadly suspicious tax behaviour can be divided in three categories: supervised, semi-supervised and unsupervised. 
Contrary to unsupervised methods, supervised and semi-supervised methods require training datasets in which previous 
cases of unwanted tax behaviour are entirely or partially indicated.

\subsection{Supervised Methods}

To the best of our knowledge, the first application of supervised machine learning techniques to the problem of fiscal fraud detection
can be found in the work of~\citet{Bonchi1999}. The authors examine C5.0 algorithm for learning decision
tress to build a classification model for detecting tax evasion in Italy. Other supervised machine learning 
approaches for detecting fraudulent tax behavior investigated in the literature include 
random forests~\citep{Mittal2018}, 
rule-based classification~\citep{Basta2009} 
and Bayesian networks~\citep{SilvaRCS16}.

A VAT screening method to identify non-compliant VAT reports for further auditing checks is proposed by~\citet{Wu2012}. The  method utilizes the Apriori algorithm for extracting association rules from a previously formed database of business entities that were involved in VAT evasion activities. The method is  evaluated on non-compliant VAT reports in Taiwan in the period 2003-2004 with 3780 confirmed VAT evasion cases. The accuracy rate of identified association rules is estimated to approximately 95\% using the 3-cross validation technique. The idea of using the Apriori algorithm for identifying a set of pattern characterizing fraudulent tax behavior is also explored by~\citet{Matos2015} with addition of dimensionality reduction techniques (Principal Component Analysis and Singular Value Decomposition) to reduce a set of fraud indicators and create a fraud scale for ranking Brazilian taxpayers. 

Both unsupervised and supervised machine learning techniques for identifying suspicious tax-related behaviour are analysed in \citep{CastellonGonzalez2013}. Two clustering techniques, self-organizing maps and neural gas, are applied to identify groups of business entities with similar tax-related behaviour in two datasets encompassing micro to small and medium to large enterprises from Chile in the period 2005-2007. Three classification learning algorithms (decision tree, neural network and Bayesian network) are then employed to construct classifiers detecting fraudulent behaviour.  The evaluation showed that the examined classification techniques are able to reach the accuracy of 92\% and 89\% for micro to small business and medium to large enterprises, respectively. The article also summarizes machine learning practices for detecting tax evasion employed by governmental agencies that were not  documented in the academic literature before.

A transfer learning method for tax evasion detection based on conditional 
adversarial networks is introduced in~\citep{Wei2019}. The  main idea is to utilize an adversarial learning to extract
tax evasion features from a labeled dataset. The trained neural network is then applied to an
unlabeled dataset. The transfer learning approach is tested on five 
tax datasets corresponding to five different Chinese regions.

\citet{Zumaya2021} analyze tax evasion on a large-scale dataset of electronic records regarding taxable transactions in Mexico by applying tools from network science and machine learning. In more detail, the authors develop network-based models and, exploring properties of network neighborhoods around known tax evaders, demonstrate that the evaders' interaction patterns differ from those of the majority of contributors. Using this insight, the authors use deep neural networks and random forests to classify other contributors as potential suspects of tax evasion.

\subsection{Semi-supervised Methods}

The problem of tax evasion detection is also approached by semi-super\-vis\-ed and weakly-supervised 
learning methods. \citet{KLEANTHOUS2020} proposed a semi-supervised method for VAT audit case selection based on 
gated mixture variational autoencoder networks. 

The method proposed by~\citet{WuZheng2019} is based on positive 
and unlabeled (PU) learning techniques. PU techniques
are applicable to datasets in which only a small subset of data instances is 
positively labeled and the rest are not annotated at all.
The method by~\citet{WuZheng2019} combines random forest to select the most relevant features, 
one-class probabilistic classification to assign pseudo-labels to unlabeled data instances 
and LightGBM as the final supervised model trained on pseudo-labeled data. 

\citet{MiDong2020} also explored PU learning for detecting tax evasion cases. The 
proposed method additionally incorporates features obtained by embedding a transaction graph 
into an Euclidean space.

The work by \citet{Gao2021} further improves the method by \citet{MiDong2020}. More concretely, the 
authors present PnCGCN -- a novel graph embedding algorithm for transaction graphs. PnCGCN is utilized to 
extract network-based features prior to assigning pseudo-labels to unlabeled data instances. In the final
stage, a multilayer perceptron neural network is trained on pseudo-labeled data to identify tax evasion 
cases.

\subsection{Unsupervised Methods}

The article by~\citet{DeRoux2018} emphasizes the importance of unsupervised machine learning methods 
for tax fraud detection due to intrinsic limitations of tax auditing processes. 
An unsupervised method for 
detecting under-reporting tax declaration based on spectral clustering is proposed. The probability distribution 
of declared tax bases is computed for each identified cluster. Suspicious tax declarations in a cluster are 
then identified using a quantile of the corresponding probability distribution. The method is  experimentally 
evaluated  on a dataset encompassing tax declarations of building projects in Bogota, Columbia.
Spectral clustering to detect tax evaders is also explored by~\citet{Priya2020}. Besides features derived from
individual tax returns, in the clustering procedure, a feature derived from a graph
showing business interactions among taxpayers is also included. This graph-based feature is computed by  
the TrustRank algorithm (a variant of the PageRank algorithm) to the constructed graph. The method is experimentally evaluated
on tax declarations in India collected in the period 2017-2019.

\citet{Tian2016} introduces graph-based techniques for mining suspicious tax evasion groups in big datasets. A method for fusing a complex heterogeneous graph called Taxpayer Interest Interacted Network (TPIIN). TPIIN is enhancing a trading graph for a set of companies by investment relationships between them, influence relationships between persons and companies, and kinship relationships between persons. Suspicious tax evasion groups are then detected as subgraphs of TPIIN consisting of simple directed trails having the same start and end nodes without any trading link between intermediate nodes belonging to different trails.

A method for detecting accounting anomalies in journal entries exported from 
ERP (Enterprise Resource Planning) information systems is presented in~\citep{Schreyer2017}. It is is based on autoencoder networks and, to the best of our knowledge, it is 
the first application of deep learning techniques to detect anomalies in large scale accounting data. The application
of autoencoders for detecting suspicious journal entities in financial statement audits is also explored by~\citet{Schultz2020}.

The article by~\citet{VANHOEYVELD2020} investigates the applications of light\-weight un\-super\-vised outlier 
detection algorithms to VAT fraud detection. The two outlier detection 
algorithms, fixed-width anomaly detection and local outlier factor, applied to VAT declarations
of ten business sectors in Belgia are analysed. The analysis of lift and hit rates of examined methods 
showed that simple and highly scalable outlier detection methods can be very effective when
identifying outliers in sectoral VAT declarations.

\subsection{Our Contributions}

Previous relevant research works clearly indicate that the applicability of supervised ML approaches to detect tax evasion is 
conditioned by training datasets containing a significant number of confirmed tax evasion cases. 
Such datasets are not always available since they result from extensive and costly tax auditing processes. 
Thus, unsupervised and semi-supervised ML techniques (when available data contains a small number of
confirmed tax evasion cases) are still dominant approaches to detect tax evasion in tax databases. 
Some of the proposed methods (in all three categories) rely on graph-based features derived from transaction graphs. 
Those methods can be applied only if data describing transactions between business entities is available.

In this paper we propose a hybrid unsupervised method to detect anomalies in tabular tax datasets without 
special restrictions or requirements on features describing business entities.
Our method combines clustering ($K$-means) and representational learning (autoencoders) to devise a set of internally validated
outliers. As the literature review shows, both clustering and autoencoders were used in previous relevant research works 
as techniques to identify outliers in tax datasets. However, none of prior research proposed
an unsupervised method for detecting internally validated outliers in tax datasets nor examined any kind of 
synergy of those two particular outlier detection approaches.

Furthermore, existing unsupervised outlier detection approaches for tax data\-sets do not incorporate domain knowledge into 
their decision models. In HUNOD, both constituent outlier detection algorithms are enhanced by relevant domain knowledge:
(1) feature weights given by an economist familiar with tax related business practices that indicate feature relevance for fraudulent tax behaviour
are incorporated into $K$-means prior to identifying clusters of business entities, and (2) the training dataset for
the autoencoder is formed by selecting a subset of business entities according to a scoring indicator reflecting
the degree of non-suspicious tax-related behaviour.

Compared with previous autoencoder-based approaches, the inclusion of domain knowledge in HUNOD and its hybrid nature 
additionally improve the robustness of the HUNOD autoencoder:
\begin{enumerate}
 \item The HUNOD autoencoder is not trained on the whole dataset, but on a subset of business entities for which we can be 
 quite confident in their regular tax-related behaviour. This has two important consequences: 
 (1) the identified latent representations are more robust since they are learned only on negative examples, and (2) 
 it is not necessary to define a threshold when deciding whether a test data instance is an outlier 
 (the maximal autoencoder error on the training dataset is a natural decision boundary). 
 \item K-means is instrumented to verify that the training dataset does not contain structural-based
 outliers (data instances located in small clusters).
\end{enumerate}

It is also important to observe differences between HUNOD and previously proposed semi-supervised approaches based 
on PU learning: the HUNOD autoencoder is trained on negative examples (regular tax behaviour), whereas PU models are 
trained on positive examples (tax evasion cases) that are less frequent than negative examples (it is natural to expect
that a model trained on bigger data would be more robust). Second, our method does not require any positive examples.

The issue of explainability of unsupervised anomaly detection techniques, although indicated in general literature as a 
big challenge~\cite{Pang2020}, was never thoroughly investigated in research works applying those techniques to tax datasets.
Our HUNOD method relies on surrogate decision trees that are able not only to provide explanations for individual outliers, 
but also to rank features according to their power to indicate outliers.

Finally, the available empirical literature on tax fraud detection mostly focuses on VAT tax 
(or similar indirect tax like sales tax used in some countries) data analysis. 
These datasets usually cover a small number of features extractable from tax declarations 
for this type of tax. In our case study we apply 
the HUNOD method to datasets of business entities in Serbia 
derived from individual personal income tax declarations.
To the best of our knowledge, there is no prior research on tax fraud detection using machine learning approaches 
to the personal income tax.

\section{The HUNOD Method}
\label{method}

The method we propose combines unsupervised clustering and representational
learning algorithms additionally incorporating relevant domain knowledge into the used unsupervised algorithms 
and enhancing the interpretability of detected outliers by training supervised, explainable-by-design 
surrogate models over the results of unsupervised learning algorithms.

An outlier is a data instance that is significantly different from vast majority of other data instances. 
Considering the clustering structure of a dataset, outliers are data instances located in small and distant clusters. The main issue with the clustering approach to outlier detection is how to set thresholds below which a cluster is considered small and distant. With reasonable upper bounds for size and 
distance thresholds, a clustering algorithm such as $K$-means yields a broad set of data points that contain all outliers, but also data points that are not outliers. In other words, a clustering-based outlier detection model can have a very high or even the perfect recall at the cost of a low precision. The second constituent component of HUNOD is an autoencoder -- a representational learning (deep) neural network. 
The HUNOD autoencoder learns latent data representation of desirable tax behavior, i.e. it is trained on data points that are the most representative examples of compliant tax behavior selected according to the domain knowledge. Contrary to $K$-means, the autoencoder-based outlier detection model is able to reach a high precision since it is a trained model. The precision of the autoencoder model can be verified and additionally improved by checking its results against the clustering model: false positive decisions made by the autoencoder can be accurately identified since the clustering model has a very high or perfect recall.

\begin{figure}[htb!]
\centering
\includegraphics[width=0.95\textwidth]{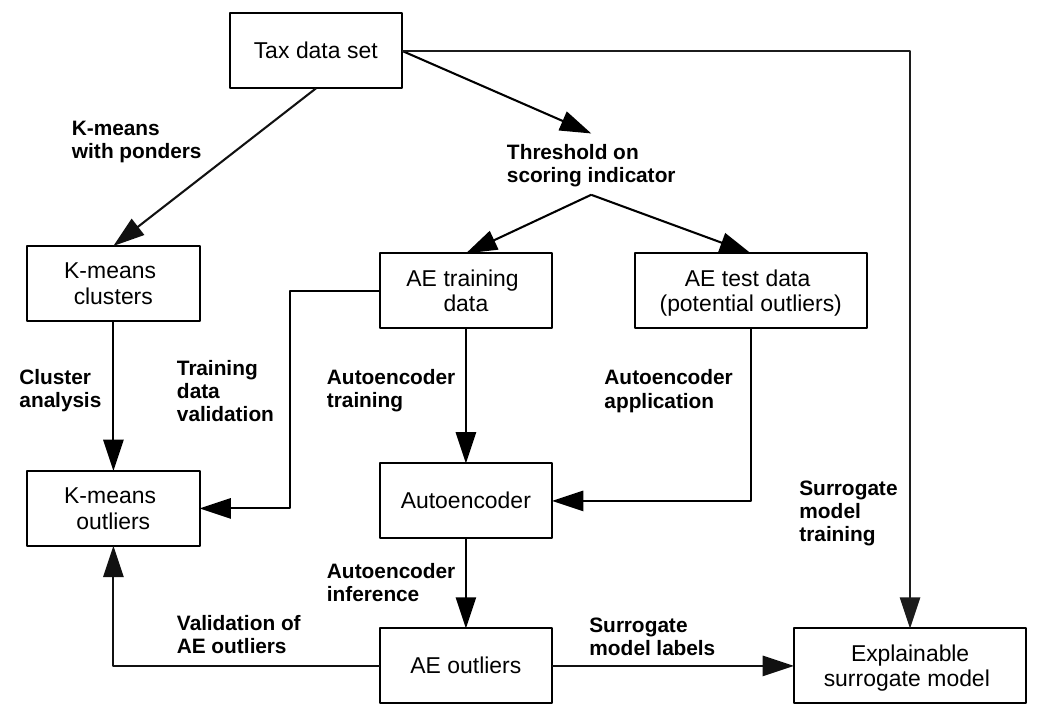}
\caption{Schematic illustration of the hybrid outlier detection method for the tax data set}
\label{fig1_method}       
\end{figure}

The constituent components of the method and the corresponding dataflow are 
schematically illustrated in Figure~\ref{fig1_method}. A modification of $K$-means incorporating 
relevant domain knowledge in terms of feature weights is used
to detect a broad set of outliers. This part of HUNOD is described in Section~\ref{K_means_with_ponders}.
A domain-knowledge scoring indicator is then utilized to identify a subset of business entities
for which we can state non-suspicious tax related behavior with a very high confidence. The instances
in this subset constitute the training dataset for an autoencoder. The training dataset for the 
autoencoder is validated by cross-checking with outliers identified by $K$-means with ponders. 
The autoencoder is then applied to all instances that are not in the training dataset to identify 
a narrower set of outliers. The set of outliers identified by the autoencoder is also verified 
against outliers detected by $K$-means with ponders. This whole process is explained in
Section~\ref{AEOD}. In the final step of HUNOD method the input dataset is extended by a binary 
feature indicating whether the corresponding instance is an outlier by the autoencoder. A decision tree
is trained on this extended dataset in order to obtain the explainable surrogate model
for detected outliers, Section~\ref{ESM}.

Throughout this section, we assume that the input tax dataset, denoted by $D$, is composed of instances 
(business entities) described by real-valued attributes or features, denoted by $f_{1}$ to $f_{d}$, 
where $d$ is the total number of features.  
This means that present categorical features are converted 
into real-valued features by employing 
the one-hot-encoding transformation prior to the application of our hybrid outlier detection method.  

\subsection{$K$-means with Ponders}
\label{K_means_with_ponders}

The hybrid outlier detection method instruments the $K$-means clustering algorithm~\citep{Lloyd_1982,macqueen1967} 
to detect a broad set of potential outliers in $D$. 
For a given $K$, the $K$-means algorithm identifies $K$ clusters in $D$ using a previously 
specified distance function (e.g., the Euclidean or Manhattan distance). The algorithm starts with
selecting $K$ random data points in the space of $D$ (i.e, a box of dimension $d$ 
bounding all instances in $D$) which serve as cluster centroids. Then, all instances in $D$ are
assigned to the closest centroid and the centroids are recomputed according to the assignments.
The previous step is repeated either a fixed number of iterations or until centroids stabilize.

The HUNOD method incorporates domain knowledge in the $K$-means clustering by introducing 
ponders or feature weights. The ponders are assigned by a tax expert according to her/his assessment 
of feature importance for outlier detection. In our realization of the $K$-means clustering with 
ponders, the distance between two instances $p$ and $q$ is computed by introducing weights to
the Euclidean distance:
\begin{equation}
\label{eqn_dist}
\mbox{distance}(p, q) = \sqrt{\sum_{i = 1}^{d}w_{i}^{2}(f'_{i}(p) - f'_{i}(q))^{2}}
\end{equation}
where $w_{i}$ is the weight of feature $f_{i}$ and $f'_{i}(r)$ is the value of feature $f_{i}$ 
for instance $r$ after feature scaling to the unit interval. The main benefit of such defined
distance function is that it enables the reduction of $K$-means with ponders to 
$K$-means without ponders by simple transformations of $D$ (see Algorithm 1). 
Consequently, existing and verified
implementations of $K$-means can be reused when realizing $K$-means with ponders. The $K$-means implementation
from the scikit-learn Python library~\citep{scikit-learn} is used in our realization.

\begin{algorithm}[htb!]
\small
\SetAlgoLined
\DontPrintSemicolon
\SetKwInOut{Input}{input}
\SetKwInOut{Output}{output}
\Input{ $D$, $W$, $S$, $K$, $M$ 
\begin{itemize}
\setlength\itemsep{-0.1em}
\item[] $D$ -- the dataset with real-valued features $f_{1}, f_{2}, \ldots, f_{d}$ 
\item[] $W$ -- the vector of feature ponders, $W = (w_{1}, w_{2}, \ldots, w_{d})$
\item[] $K$ -- the number of clusters
\item[] $S$ -- the threshold separating large from small clusters
\item[] $M$ -- the maximal number of outliers returned by the algorithm
\end{itemize}
}
\Output {$O$ -- the set of outliers ($O \subset D$)} 

\BlankLine \BlankLine 
{\bf // Phase 1 -- clustering}\\
scale features $f_{1}, f_{2}, \ldots, f_{d}$ to the unit interval to 
obtain dataset $D'$ with features $f'_{1}, f'_{2}, \ldots, f'_{d}$ 
\BlankLine
\For{$j = 1 \: \mbox{to} \: d$} {
    \ForEach{$i \in D'$} {
        $f'_{j}(i) = f'_{j}(i) * w_{k}$
    }
}
\BlankLine 
$C$ = apply $K$-means to $D'$ using the Euclidean distance to obtain $K$ clusters\\ 

\BlankLine \BlankLine 
{\bf // Phase 2 -- cluster analysis} \\
$C_{s} = \{c \in C: \: |c| < S\}$   // small clusters\\ 
$C_{l} = \{c \in C: \: |c| \geq S\}$   // large clusters\\
$l$ = an empty list of tuples (cluster, outlierness score) \\
\ForEach{$c \in C_{s}$} {
    // compute the outlierness score for cluster $c$\\
    $s$ = distance from the centroid of $c$ to the closest centroid of clusters in $C_{l}$\\
    append ($c$, $s$) to $l$
}

\BlankLine
sort $l$ in descending order of the outlierness score\\
$O = \emptyset$ \\
$i = 1$ \\
$w = \top$ \\
\While{$(i \leq \: \textup{length of} \: l) \:\: \wedge \: w$} {
    $C$ = $i$-th cluster in $l$\\
    \uIf{$|C| + |O| \leq M$} {
        $O = O \cup C$
    }\Else{
        $w = \bot$
    }
}

{\bf return} $O$

\caption{{\bf Outlier detection based on $K$-means with ponders}}	
\SetAlgoRefName{Al1}
\SetAlgoCaptionSeparator{'.'}
\end{algorithm}

Clusters identified by $K$-means with ponders are then analyzed to identify 
outliers. In the first step, clusters are classified either as
small or large, depending on the value of parameter $S$: a cluster
having less than $S$ instances is considered as small, otherwise it is large. 
The default value of $S$ is  5\% of the total number of instances in $D$.
Clearly, potential outliers are located in small clusters that are  
far away from large clusters. Thus, for each small cluster we 
compute the outlierness score according to the following 
formula:
\begin{equation}
\label{outlierness}
\mbox{outlierness}(s) = \min_{c \in C_{l}} \: \{\mbox{distance}(\mbox{centroid}(s), \mbox{centroid}(c))\}
\end{equation}
where $C_{l}$ is the set of large clusters and the distance between centroids
is computed by~(\ref{eqn_dist}).
$K$-means with ponders also has a parameter $M$ which is the maximal number of outliers returned
by the algorithm. The default value of this parameter is 5\% of the total number of 
instances in $D$. The small clusters are sorted in descending order 
of the outlierness score. The instances from the first $k$ clusters 
in the sorted sequence are 
selected as outliers, where $k$ is the largest index such that the total
number of instances in the first $k$ clusters is smaller than or equal to 
$M$.

\subsection{Autoencoder-based Outlier Detection}
\label{AEOD}

The central part of HUNOD method is the training and application of an autoencoder 
outlier detection. The autoencoder is a feed-forward neural network trained to reconstruct values of 
an input set of features at the output layer through a series of hidden layers. 
This implies that the number of neurons in the input layer
of the autoencoder is equal to the number of neurons in its output layer.
Each hidden layer contains a specified number of nodes (neurons). 
Each node accumulates inputs from all nodes in the previous layer that are 
multiplied by weights of edges connecting nodes, 
adds a bias value to the accumulated 
input and applies an activation function to the accumulated input to produce 
its output. Edge weights and biases are trainable model parameters.
The first $m$ layers of the autoencoder represent an encoding function 
transforming an instance (input values) into a lower-dimensional representation, 
while the rest of the layers act as a decoding function transforming the obtained 
lower-dimensional encoding into original input values.

In common outlier detection applications, the autoencoder is trained on all instances
in a given dataset, and the instances with the highest reconstruction error are
selected as outliers~\citep{hawkins_2002}. In our approach, the autoencoder is trained
on a subset of instances in $D$ selected according to a scoring indicator
derived from  domain knowledge. The scoring indicator is a function computed
on input features (or a subset of them) indicating the degree of non-suspicious 
tax related behaviour of the corresponding business entity. 
The details on the scoring indicators we use are presented in~\citep{nas}.  
The hybrid outlier detection method
trains the autoencoder on instances with high values of the scoring indicator
(values above a threshold that is assigned by a tax expert). In other words,
the autoencoder is trained on those business entities for which there is a 
high confidence that their tax related behaviour is not suspicious. Additionally,
the accuracy of the training dataset is estimated by cross-checking with 
outliers detected by $K$-means with ponders motivated by the fact that
instances in the training dataset should not be outliers detected by
some other outlier detection method.

Let $T$ denote a subset of $D$ encompassing instances whose values of the scoring 
indicator are above the specified threshold,
$L$ the number of autoencoder layers (indexed from 1 to $L$), 
$x^{(l)}$ the vector of input values for layer $l$ and $y^{(l)}$ the vector of output 
values from layer $l$. 
The feed-forward operation of the autoencoder for an instance $p$ from $D$ 
then can be recursively defined as
\begin{equation}
\label{autoencoder_feed_forward}
\begin{array}{r@{}l}
    x_{i}^{(l)}(p) &{} = W_{i}^{(l)}y^{(l - 1)}(p) + b_{i}^{(l)}\\
    y_{i}^{(l)}(p) &{} = \sigma(x_{i}^{(l)}(p))\\
    y_{i}^{(0)}(p) &{} = f_{i}(p)
\end{array}
\end{equation}
for each hidden node $i$ and layer $l$. As above, $f_{i}(p)$ is the value of $i$-th feature
of the instance  $p$, while $\sigma$ is a non-linear activation function, e.g. 
sigmoid $\sigma(x) = 1 / (1 + \mbox{exp}(-x))$ or rectified linear unit (ReLU) 
$\sigma(x) = \mbox{max}(0, x)$. The weights of autoencoder edges $W$ and biases $b$
are learned by minimizing a loss function $\Loss$ considering instances in $T$.
HUNOD method uses the mean squared error (MSE) for the loss function
\begin{equation}
\label{loss_function}
\Loss(T) = \frac{1}{|T|} \: \sum_{p \in T} \: E(p)
\end{equation}
where $E(p)$ is the reconstruction error of instance $p$
\begin{equation}
\label{error_function}
E(p) = \sum_{i = 1}^{d} (y_{i}^{L}(p) - f_{i}(p))^{2}
\end{equation}

The trained autoencoder is then applied to all instances in $D \setminus T$.
An instance $q \in D \setminus T$ is marked as an outlier if the reconstruction
error for $q$ is higher than the maximal reconstruction error on the training 
dataset $T$
\begin{equation}
\label{ae_outlier}
E(q) > \max_{p \in T} \: \{E(p)\} 
\end{equation}
Moreover, the reconstruction error $E(q)$ can be viewed as the outlierness score for
$q$. All outliers detected by the autoencoder are then verified against outliers
identified by $K$-means with ponders. 

To prevent overfitting we include also dropout~\citep{dropout_2014} and 
$L_2$ activation regularization~\citep{activityreg_2018} 
mechanisms into autoencoder training. Both mechanisms increase the sparsity of the autoencoder:
the dropout by ignoring randomly selected neurons during training and the $L_2$ activation regularization
by introducing an activation penalties into the loss function. With the dropout included 
equations~(\ref{autoencoder_feed_forward}) change to
\begin{equation}
\label{autoencoder_feed_forward_dropout}
\begin{array}{r@{}l}
    r_{j}^{(l)}    &{} \sim \mbox{Bernoulli}(\alpha)\\
    x_{i}^{(l)}(p) &{} = W_{i}^{(l)} (r^{(l)} \circ y^{(l - 1)}(p)) + b_{i}^{(l)}\\
    y_{i}^{(l)}(p) &{} = \sigma(x_{i}^{(l)}(p))\\
    y_{i}^{(0)}(p) &{} = f_{i}(p)
\end{array}
\end{equation}
where $j$ denotes any autoencoder edge, $r^{(l)}$ is a vector of independent Bernoulli random variables at 
layer $l$, $\alpha$ is the probability that an edge is not  discarded ($0 \leq \alpha \leq 1$) and 
$\circ$ represents the element-wise product operation. With the $L_2$ activation regularization, the loss 
function~(\ref{loss_function}) becomes
\begin{equation}
\label{loss_function_actreg}
\Loss'(T) = \frac{1}{|T|} \: \sum_{p \in T} \: E(p) \: \: + \: \: \lambda \sum_{p \in T}\sum_{l = 1}^{L}\sum_{n \in N(l)} y_{n}^{(l)}(p)^{2}
\end{equation}
where $N(l)$ denotes the set of neurons in layer $l$ and $\lambda \geq 0$ is the regularization factor. 

In HUNOD method, the Tensorflow framework~\citep{Tensorflow_2016} 
is used for training the autoencoder and instrumenting it to perform outlier detection inference. 
The whole procedure is shown in Algorithm 2. The autoencoder is built from the Tensorflow sequential
neural network model by adding the specified number of hidden layers. The default values of
the regularization hyperparameters are $\alpha = 0.8$ and $\lambda = 0.1$. The $L_2$ activity 
regularization can be excluded by setting $\lambda$ to 0. The loss function is optimized 
using the Adam optimization algorithm~\citep{Adam_2015} in a given number of epochs $e$ (the number of forward and
backward passes of $T$ through the autoencoder to optimize model parameters -- edge weights and biases) and 
batch size $b$ (the number of instances processed to update model parameters). 
The default values of the autoencoder hyperparameters are $e = 200$ and $b = 32$. The activition function is set to ReLU, $\sigma(x) = \mbox{max}(0, x).$

\begin{algorithm}[htb!]
\small
\SetAlgoLined
\DontPrintSemicolon
\SetKwInOut{Input}{input}
\SetKwInOut{Output}{output}
\Input{ $D$, $S$, $S_{t}$, $L$, $N$, $\alpha$, $\lambda$, $O$
\begin{itemize}
\setlength\itemsep{-0.1em}
\item[] $D$ -- the dataset with real-valued features $f_{1}, f_{2}, \ldots, f_{d}$ 
\item[] $S$ -- the scoring function
\item[] $t$ -- the threshold for selecting training instances from $D$
\item[] $L$ -- the number of hidden layers
\item[] $N$ -- the array containing the number of neurons per hidden layer
\item[] $\alpha$, $\lambda$ -- regularization hyperparameters
\item[] $e$, $b$ -- autoencoder hyperparameters (epochs and batch size)
\item[] $O$ -- the set of outliers detected by $K$-means with ponders
\end{itemize}
}
\Output {$A$, $C_t$, $C_a$
\begin{itemize}
\setlength\itemsep{-0.1em}
\item[] $A$ -- the set of outliers detected by the autoencoder
\item[] $C_t$ -- the confidence in the training dataset estimated using $O$ 
\item[] $C_a$ -- the confidence in $A$ estimated using $O$
\end{itemize}
} 

\BlankLine \BlankLine 
{\bf // Phase 1 -- training data preparation}\\
$T = \{i \in T \: : \: S(i) \geq t\}$\\
$C_t = |T \cap O| / |T \cup O|$\\
$P = D \setminus T$\\

\BlankLine \BlankLine 
{\bf // Phase 2 -- autoencoder training}\\
ae = Sequential()\\
ae.addLayer(Dense(neurons=$d$, activation=ReLU, reg=$L_2$($\lambda$)))\\
\For{$c = 1 \: \textup{to} \: L$} {
    ae.addLayer(Dense(neurons=$N[i]$, activation=ReLU, reg=$L_2$($\lambda$)))\\
    \If{$\alpha \: \neq \:$ \textup{1}} {
        ae.addLayer(Dropout(rate=1-$\alpha$))
    }
}
ae.addLayer(Dense(neurons=$d$, activation=Linear))\\
ae.compile(optimizer=Adam, loss=MSE)\\
ae.fit(predictors=$T$, target=$T$, epochs=$e$, batchSize=$b$)\\
$M = \max_{p \in T} \: \{ \mbox{ae.inference}(p)\}$\\ 

\BlankLine \BlankLine 
{\bf // Phase 2 -- autoencoder application}\\
$A = \{ p \in P \: : \: \mbox{ae.inference}(p) > M\}$ \\
$C_a = |A \cap O| / |A \cup O|$

\BlankLine \BlankLine
{\bf return} $A$, $C_t$, $C_a$ 
\caption{{\bf Autoencoder-based outlier detection}}	
\SetAlgoRefName{A$L_2$}
\SetAlgoCaptionSeparator{'.'}
\end{algorithm}

\subsection{Explainable Surrogate Model}
\label{ESM}

In the final step of the hybrid outlier detection method, outliers detected by
the autoencoder (and cross-checked against $K$-means with ponders) are used to
extend the input dataset $D$ by an extra binary class feature $C$. All instances
identified as outliers are marked to belong to the positive class ($C = 1$), while
other instances are assigned to the negative class ($C = 0$). An explainable-by-design 
surrogate model performing binary classification is then trained on the extended dataset. 
The design of the surrogate model shows 
common characteristics of outliers 
identified by the autoencoder in terms of feature values. Additionally, 
the surrogate model enables us to rank features according to their 
discriminative power to separate outliers from non-outliers.

Classic decision trees~\citep{Quinlan_1986} are instrumented in our work as 
explainable surrogate 
models for outliers identified by the autoencoder. Each non-leaf node in the 
binary decision tree is a relational expression involving one feature and a constant
value (e.g., $a > 10$ where $a$ denotes a real-valued feature). Two edges
emanates from each non-leaf node: the left edge is activated if the
relational expression evaluates to false, otherwise the right edge is activated.
The activated edge leads either to a leaf node or some other non-leaf node.
Leaf nodes of the decision tree represent classes. A path from the root node
to a leaf node through activated edges for a given input instance then 
explains how it is classified and which criteria it satisfies to be assigned
to the class represented by the leaf node. In a $n$-ary decision tree,
the nodes represent features, while edges are relational expressions. Clearly,
each $n$-ary decision tree can be converted to an equivalent binary 
decision tree. 

A decision tree for a given labeled dataset $E$ can be constructed by
a recursive divide-and-conquer algorithm guided by a splitting function.
For a feature $f$ and a relational expression $r$ involving $f$, 
the splitting function measures the extent to which the splitting of $E$ by 
$r$ produces ``pure'' splits in terms of class frequencies (a split
is pure if it contains only instances belonging to the same class). 
The splitting function utilizes a function measuring the purity of
a dataset. Two commonly used approaches are the Gini impurity and
the entropy~\citep{Suthaharan2016}. In the binary classification setting (which is the case 
for our surrogate models), those two measures are defined as
\begin{equation}
\label{purity_functions}
\begin{array}{r@{}l}
    \mbox{Gini}(E)       &{} = C^{+}(1 - C^{+}) + C^{-}(1 - C^{-})\\
    \mbox{Entropy}(E)    &{} = -C^{+}\mbox{log}(C^{+}) -C^{-}\mbox{log}(C^{-}) \\
\end{array}
\end{equation}
where $C^{+}$ and $C^{-}$ are the fractions of positive and negative instances
in $E$, respectively. Clearly, Gini($E$) = 0 or Entropy($E$) = 0 implies that
all instances in $E$ are from the same class (either positive or negative). 
Secondly, smaller values of both measures indicate a higher degree of purity.

The splitting function measures the change in the impurity of $Q$ after splitting
by $r$, where $Q$ is $E$ or a subset of $E$ obtained after applying one or more
splitting operations:
\begin{equation}
\label{splitting_functions}
S(Q, r) = L(Q)P(Q) - L(Q_1)P(Q_1) - L(Q_2)P(Q_2)
\end{equation}
where $P$ is a dataset impurity measure (e.g., Gini or Entropy), $Q_{1}$ are instances
from $Q$ not satisfying $r$, $Q_{2} = Q \setminus Q_{1}$ and $L(R)$ is the fraction
of instances in split $R$, $R \in \{Q, Q_1, Q_2\}$, $L(R) = |R| / |E|$. 
Now, an algorithm Binary Decision Three,  BDT($E$) for building the decision tree on 
$E$ can be formulated as follows:
\begin{itemize}
 \item If $E$ contains only one feature or all instances in $E$ belong to the same class
 then return a leaf node with the dominant class in $E$.
 \item Otherwise, find a relational expression $r$ leading to the highest $S(E, r)$.
 \item Split $E$ according to $r$ to obtain $E_{1}$ and $E_{2}$. 
 \item Make a node $R$ in the decision tree corresponding to $r$.
 \item Make recursive calls BDT($E_{1}$) and BDT($E_{2}$) to create the left and right
 subtree for $R$, respectively.
\end{itemize}
Since $S(Q, r)$ is computed for each node in the decision tree, the importance of feature 
$f$ for the classification process imposed by the decision tree can be defined as the 
sum of $S$ values of nodes associated to $f$ normalized by the sum of $S$ values for all
nodes in the tree.

The implementation of the decision learning algorithm from the Scikit-learn library~\citep{scikit-learn}
is used in the HUNOD method to learn explainable surrogate models. 
The HUNOD method executes the Scikit-learn decision tree learning algorithm using 
its default parameters as given in the Scikit-learn library: 
(1) the best splits are determined by the Gini impurity measure, 
(2) the number of internal and leaf nodes and the maximal depth of the resulting decision tree are not 
bounded to some predefined values, 
and (3) the minimal number of instances to perform splitting is equal to 2.
The Scikit-learn library is additionally utilized to evaluate the surrogate model on the training data
by computing the accuracy score. The accuracy score is the number of correct classifications divided by 
the total number of instances on which the classification process is performed. Clearly,
a reliable surrogate model must have the accuracy score on the training dataset equal
to 1. In other words, the trained surrogate model has to be able to predict the same 
outliers as the autoencoder prior to providing explanations 
for particular outliers in terms of feature values.

\section{Experimental Dataset}
\label{dataset}
The experimental dataset is derived from the database on individual personal income tax declarations collected by the Tax Administration of Serbia. According to the ruling legal framework, the individual tax declaration for personal income is typically submitted by the payer of income. There is about 0.1\% of all declarations relating to 0.001\% of individuals in the observed data where an individual which is a receiver of income has submitted the tax declaration (e.g. when income is received from abroad) and these cases are not included in our experimental dataset. An individual tax declaration consists of data on payer of income ID (tax identification number, TIN), receiver of income ID (personal identification number, PID), date of  income payment, date of declaration, date of payment of tax and contributions, type of income, gross amount of income, amount of paid tax and mandatory contributions for health insurance, pension insurance and unemployment insurance. Type of income refer to different types of personal income that are subject to personal income tax and mandatory contributions for social insurance. These types of income are: salary, sick leave compensations, dividend, interest, rent, authors income, temporary work assignment and one-off performance service contract. Additionally, the data on individual income payer (TIN) is accompanied with data on its sector of prevalent economic activity (NACE code), location of headquarter, date of establishment. Moreover, an individual income receiver is described by gender, age and its residency location municipality code. 

In order to investigate tax fraud which is predominantly the activity performed by payer of income, the original dataset is transformed to show the business entity (TIN) level aggregated attributes on monthly level. In our experimental dataset, we observe all different TINs appearing in tax declarations along 13 months period (from March 2016 to March 2017). We excluded from the dataset public administration organizations, some non profit associations and one large state energy company which is a public monopoly. In this way we obtained the empirical dataset consisting of ca. 22 
million data fields.

The final dataset covers 179,489 different business entities (TINs) which are mostly enterprises. We observe 224 different features for each TIN. Three features describe TINs status and don’t change over time. These are industry code and organization form which are represented by as many binary variables as there is different values (422 different industry codes and 24 different organizational forms) as well as the age of TIN being the difference between the year of establishment and the observed year. Other features are derived as different statistics calculated on the TIN level using original tax declaration data for each month. As these values are changing over time, each of thirteen months of the observed period represent a distinct feature. Therefore the 221 features can be observed as 17 differently defined features using the individual tax declaration data submitted by one TIN in one month each appearing as 13 different features denoted by adding {\it YmN} in the name of the feature, where {\it Y} is year and {\it N} is month number. These 17 different feature definitions aim to capture some of the relevant characteristics of TINs. 

The definition of these features, though limited by the availability of data in original tax declaration, is largely motivated by the existence of some background theory so that the features can represent a good foundation for capturing risky behavior in terms of tax avoidance and tax evasion. 

We can observe these 17 variables under the four categories: characteristics of the distribution of paid salaries, capital vs. labor income, overall tax and fiscal burden and characteristics of individuals that are paid by a business entity (TIN) either as employees (for which salary is paid) or as other type of contracts generating personal income. As the features may have somewhat different meaning for small business entities (e.g., micro enterprises) 
and large business enterprises, we divide the database into two subsets with a cutoff at 10 employees. 

\subsection{Salary Distribution Features}

For the characteristics of the distribution of paid salaries within one business entity, we use mean, median and standard deviation of paid salary by a business entity in one month (features named {\it average salary YmN}, {\it median salary YmN}, {\it stdev salary YmN}). The rationale behind including these features is that market forces and characteristics of specific business (like technology and other industry specifics) should result in convergence of the composition and a level of salaries at the level of a business entity. Therefore, identified anomalies could indicate undeclared workers and/or underreported income of declared employees. 

Next feature incorporating a domain knowledge is also related to the shape of the entity level distribution of salaries. It aims to capture the possibility for tax arbitrage and reduction of the overall fiscal burden on the business entity level. The salaries are divided into 26 bins, each bin with 10\% of increase. The tax rules contain a  cap on basis for mandatory social insurance (health, pension and unemployment) defined as five time average salary for the period and that amount is  in the 26th bin. Thus  a firm can use this rule as an opportunity to declare a few well paid individual salaries and to use these large but properly taxed salaries  for cash payment of salaries of undeclared workers. We define this feature (named {\it fraction b26r in YmN}) as follows 
$$ \frac{S_{25}-O_{26}}{S_{25}+O_{26}}, $$
where $ S_{25} $ is the sum of salaries below the defined threshold, in our case sum of the salaries in the first 25 bins of the distribution, and $ O_{26} $ is the sum of salaries in the 26th bin of the distribution, i.e. the sum of excessively large salaries. 


\subsection{Capital-Labor Ratio Features}

Captial-labor featurs are based on the ratio 
$d/S$,
where $d$ are dividends paid to owners, or other type of owners income resulting from profit, while $S$ is the sum of paid salaries and paid profit by the same business entity in one month (feature named {\it capital labor YmN}). The economic rationale behind this feature is to capture the extensive tax arbitrage given the fact that there is a significant difference in overall fiscal burden on gross salary and paid dividend to owners. Therefore, firm owners sometimes opt for paying extensive profit (after paying a relatively low corporate income tax) and use the money to pay undeclared employees in cash. As profit is usually not distributed by regular monthly payments, unlike labor income i.e. salaries, one additional feature is calculated on the aggregated yearly level (feature {\it capital labor 12m}). On the other hand, one would expect that firms operating in a relatively competitive market and using a similar technology, should register the same capital/labor ratio which is optimal in terms of profit maximization (as described in economic theory like Solow’s growth model~\citep{Solow56}).

\subsection{Tax and Fiscal Burden Features}

We defined three features to capture the overall effect of a potential underreporting of income by a business entity or a tax arbitrage: fiscal burden of salaries {\it fbs YmN}, tax burden of salaries {\it tbs YmN},  and overall fiscal burden of all type of income, {\it fball YmN}. The fiscal burden of salaries is calculated as the ratio of the sum of all paid duties by a business entity including tax and social contributions  and the sum of gross salaries paid in the same month. This feature, also known as tax wedge, is perceived as particularly high and burdensome by employers~\citep{Schneider2015} and represent an important incentive for tax evasion with the aim to lower it. Tax burden of salaries is calculated in the same way as a previous feature but  includes only paid taxes in numerator. It is worth mentioning that there is much less diversity in terms of taxing of salaries than in terms of obligatory social security regimes and therefore there is a logic to include this feature. The third feature of this type refer to overall fiscal burden on the business entity level. It is computed as 
$ {T}/{I}, $ where $ T $ is the sum of  all paid taxes and contributions by a business entity for all type of paid personal income in one month and $ I $ is the sum  of related gross income in the same month. This last indicator possibly captures if there is a tax arbitrage between various type of income. As already mentioned in Introduction, there is a complex and diverse set of taxation regimes with uneven rates across different types of income~\citep{Schneider2015}. Although some tax arbitrage can be considered as a rational economic behavior and in line with the legal framework, an extremely low fiscal burden on the business entity level can be a reflection of an excessive arbitrage or tax evasion and can signal a tax related risky behavior by business entity.  

\subsection{Features Reflecting Structure of Income Receivers}

The last category of features include: number of different individuals for which an entity pays salary in a given month ({\it total employees in YmN}), number of different individuals for which an entity pays any kind of income in a given month ({\it total persons in YmN}), average age of all individuals for which an entity pays any kind of income in a given month (feature named {\it avg age in YmN}) and seven similarly designed features capturing average age of different individual paid by the same business entity in a given month for different income categories. These are salaries, sick leave compensations, owners income, service contract fee, rent, author contract fee and other types of income (features named {\it avg age salary in YmN}, {\it avg age sick leave in YmN}, {\it avg age service fee in YmN}, {\it avg age rent in YmN}, {\it avg age owner income in YmN}, {\it avg age author in YmN}, {\it avg age other in YmN}).

The last set of features is interesting from tax fraud detection perspective in terms of its dynamics through 13 distinct features -- one for each consecutive month of the observed period. The idea is to capture the unusual dynamics (e.g. sparce payments) which can signal a potentially fraudulent behavior.  Moreover, the feature representing the average age of employees is expected to be related to the level of paid salaries (captured by the first group of features) as it is a proxy for job experience being one of two main determinants of the earning equation as evidenced by the relevant empirical and theoretical literature~\citep{Jacob74,Heckman2003}.

\section{Results and Discussion}
\label{results}

The HUNOD method is  evaluated on two independent
subsets of the experimental dataset. The first subset, denoted by $L_{10}$,
encompasses business entities with less than 10 employees. The second subset,
denoted by $A_{10}$, contains business entities having 10 or more employees.
The number of instances in $L_{10}$ and $A_{10}$ are 159744 (89.57\% 
of the total number of instances) and 18611 (10.43\%), 
respectively. 

In our experiments, $K$-means with ponders is  applied using
the following pondering scheme given by a tax expert: all features
associated to fiscal burden on paid gross salaries have weights equal to 10, 
a factor of 5 is given to all features associated to salaries and 
overall fiscal burden on paid gross income, while weights
of other features are equal to 1. 

The average of all thirteen overall fiscal burden on paid gross income ({\it fball YmN}) by business entity is instrumented
as the scoring indicator to form the training dataset for the autoencoder
for both $L_{10}$ and $A_{10}$. Business entities with the average 
overall fiscal burden on paid gross income higher than 30\% of the paid gross 
income can be considered 
non-suspicious with respect to their tax behavior and such entities
are selected as training instances. The total number of training instances
in $L_{10}$ is 31533 (19.74\% of $L_{10}$ size),
while the total number of training instances in $A_{10}$ is equal to
267 (1.43\% of $A_{10}$ size).

\subsection{Outliers by $K$-means with Ponders} \label{6.1}

The first step in our experimental analysis of the hybrid outlier detection
method is the evaluation of $K$-means with ponders.
The cluster structure of $L_{10}$ and $A_{10}$ datasets is examined 
for K taking values in \{10, 15, 20, 25, 30\}. For both datasets,
the threshold for separating small from large clusters is
set to 5\% of the total number of instances. The maximal number of
outliers returned by the algorithm is 1\% for $L_{10}$ (larger
dataset containing 159744 instances) and 5\% for $A_{10}$ 
(smaller dataset containing 18611 instances).

Characteristics of small clusters identified by $K$-means with
ponders are summarized in Table~\ref{T1}. It can be observed 
that a majority of clusters, except for $A_{10}$ when $ K $ is equal to 10,
are actually small clusters. Additionally, the number of instances
in small clusters increases with K. Table~\ref{T1} also shows 
the number of small clusters containing outliers and the fraction 
of outliers for previously mentioned thresholds limiting 
the maximal number of outliers. All outliers detected in $L_{10}$
belong exactly to one small cluster for all examined $K$ values. 
The outlierness score of the cluster containing outliers
varies between 11.68 and 11.93. The second largest outlierness
score is less than 6 implying that $L_{10}$ has a small cluster
very distant from other large clusters. Contrary to $L_{10}$, 
outliers in $A_{10}$ are  contained in more than one 
cluster for $ K > 10$.

\begin{table}[ht]
\footnotesize
\centering
\caption{Clusters identified by $K$-means with ponders. SC -- the number of small clusters,
SCS -- the percentage of instances in small clusters, SCO -- the number of small clusters
containing outliers.}
\begin{tabular}{lllllllll}
\noalign{\smallskip}\hline \noalign{\smallskip}
  & \multicolumn{4}{l}{$L_{10}$ dataset} & \multicolumn{4}{l}{$A_{10}$ dataset} \\
\noalign{\smallskip}\hline \noalign{\smallskip}
K  & SC & SCS [\%] & SCO & Outliers [\%] &  SC & SCS [\%] & SCO & Outliers [\%] \\
\noalign{\smallskip}\hline \noalign{\smallskip}
10 & 7 & 23.68 & 1 & 0.84 & 4 & 11.44 & 1 & 3.74  \\
15 & 13 & 44.16 & 1 & 0.83 & 8 & 19.73 & 2 & 4.65  \\
20 & 17 & 45.74 & 1 & 0.82 & 14 & 30.42 & 2 & 4.39 \\
25 & 22 & 55.43 & 1 & 0.81 & 21 & 43.58 & 2 & 3.49 \\
30 & 28 & 59.86 & 1 & 0.81 & 25 & 44.96 & 2 & 3.58 \\
\noalign{\smallskip}\hline \noalign{\smallskip}
\end{tabular}
\label{T1}
\end{table}

\begin{figure}[ht]
\centering
\begin{subfigure}{.5\textwidth}
  \centering
  \includegraphics[width=1\linewidth]{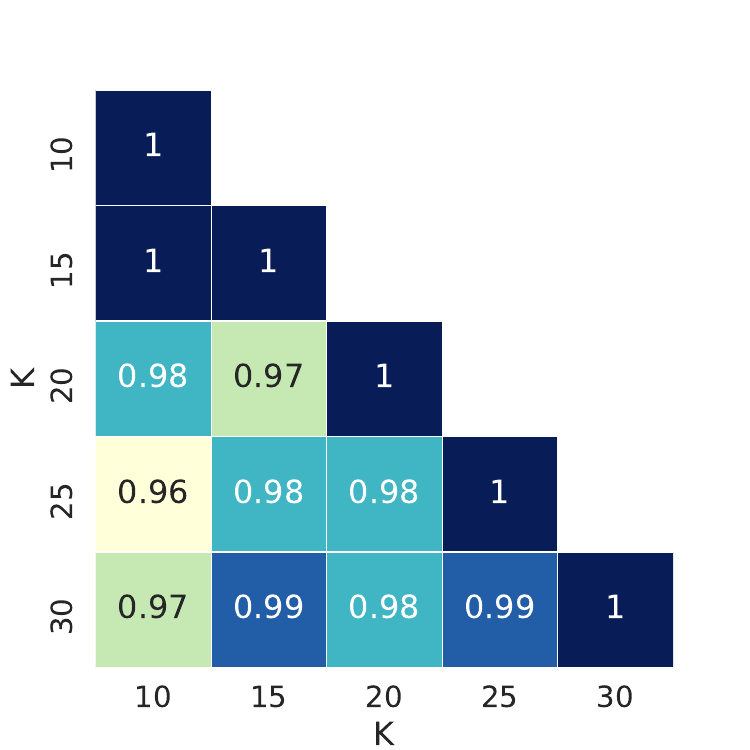}
  \caption{$L_{10}$ dataset}
  \label{fig1:sub1}
\end{subfigure}%
\begin{subfigure}{.5\textwidth}
  \centering
  \includegraphics[width=1\linewidth]{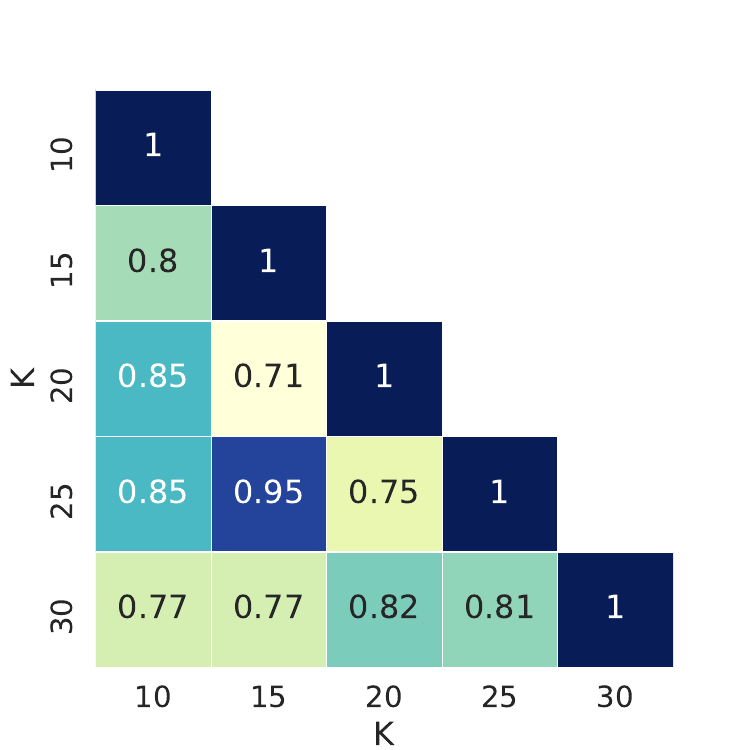}
  \caption{$A_{10}$ dataset}
  \label{fig1:sub2}
\end{subfigure}
\caption{The overlap between outlier sets for different K values measured
by the Jaccard coefficient.}
\label{fig1_jaccard}
\end{figure}

To compare outliers obtained for different $ K$ values we compute the
Jaccard coefficient for each pair of results. For two sets of 
results $A$ and $B$ this coefficient is defined as
\begin{equation}
\label{jaccard}
\mbox{jaccard-coefficient}(A, B) = \frac{|A \cap B|}{|A \cup B|}
\end{equation}
The obtained values of the Jaccard coefficient are shown in 
Figure~\ref{fig1_jaccard}. It can be seen that $K$-means with ponders applied on $L_{10}$ identifies almost identical outliers for
different $K$ values. The minimal value of the Jaccard coefficient for
$A_{10}$ is 0.71 implying that $K$-means with ponders exhibits a 
high degree of robustness to variations of the parameter $K$. 

\subsection{Outliers by Autoencoder and Surrogate Model}
\label{autoencoder_outliers}

In the second experiment we examine outliers identified by the autoencoder
and the accompanying surrogate models. The autoencoder is configured to the following
parameters:
\begin{itemize}
 \item three hidden layers containing consecutively $d/2$, $d/4$ and $d/2$ neurons ($d$ denotes the total 
 number of features),
 \item the number of epochs is 200 with the batch size of 32 instances, and
 \item the dropout probability is set to 0.2 and the $L_2$ regularization parameter $ \lambda = 0.1. $
\end{itemize}
The training dataset for the autoencoder and outliers identified by it are cross-checked
against outliers detected by $K$-means for various $ K $ values (the same values as in the 
 experiment described in part \ref{6.1}).

The total number of outliers identified by the autoencoder in $L_{10}$ dataset is
155, which is 0.097\% of the total number of instances in $L_{10}$. For $A_{10}$ dataset,
the number of outliers is equal to 137 (0.736\% of the instances in $A_{10}$). Thus, the first
conclusion that can be derived from this result is that the autoencoder identifies a significantly
smaller number of outliers than $K$-means with ponders. Let $O_{K}$ denotes a set of
outliers revealed by $K$-means with ponders for a fixed $K$. Table~\ref{T2} summarizes the results
of the cross-check against $O_{K}$ showing two quantities: 
\begin{enumerate}
 \item $V_{T}$ -- the fraction of autoencoder training instances that are in $O_{K}$, and
 \item $V_{O}$ -- the fraction of outliers detected by the autoencoder that belong to $O_{K}$.
\end{enumerate}
In case of $L_{10}$ dataset, less than 5.5\% of autoencoder training instances are in $O_{K}$ 
for various K values. This fraction can be considered small, especially 
having in mind that approximately 20\% instances in $L_{10}$ constitute the 
training dataset for the autoencoder. For $A_{10}$ dataset, we have that less than 2.6\% 
of instances in the dataset for training the autoencoder are $K$-means outliers. Thus, it 
can be concluded that the training dataset in both cases contains a negligible number
of instances that were identified as outliers by $K$-means with ponders. This
further implies that the training dataset for the autoencoder, formed
according the criterion coming from the domain knowledge, can be considered 
valid due to a high degree of consistency with an independent
outlier detection approach.

\begin{table}[ht]
\centering
\caption{Autoencoder training dataset and outliers identified by the autoencoder cross-checked
against $K$-means with ponders for different $K$ values.}
\begin{tabular}{lllll}
\noalign{\smallskip}\hline \noalign{\smallskip}
  & \multicolumn{2}{l}{$L_{10}$ dataset} & \multicolumn{2}{l}{$A_{10}$ dataset} \\
\noalign{\smallskip}\hline \noalign{\smallskip}
K  & $V_{T}$ &  $V_{O}$ &  $V_{T}$ & $V_{O}$ \\
\noalign{\smallskip}\hline \noalign{\smallskip}
10 & 0.053 & 0.987 & 0.015 & 0.908 \\
15 & 0.041 & 0.987 & 0.026 & 0.931 \\
20 & 0.055 & 0.987 & 0.022 & 0.931 \\
25 & 0.046 & 0.987 & 0.022 & 0.908 \\
30 & 0.041 & 0.987 & 0.022 & 0.915 \\
\noalign{\smallskip}\hline \noalign{\smallskip}
\end{tabular}
\label{T2}
\end{table}

Regarding the validation of autoencoder based outliers by $K$-means with ponders, it can be 
observed that
\begin{enumerate}
 \item 98.7\% of autoencoder based outliers are in $O_{K}$ for $L_{10}$ dataset, and
 \item more than 90\% of autoencoder based outliers are in $O_{K}$ for $A_{10}$ dataset. 
\end{enumerate}
In other words, a very large fraction of autoencoder based outliers are actually outliers
identified by both  independent outlier detection approaches that have nothing in common (one 
approach based on clustering, another on representational learning).
Since $K$-means with ponders identifies a significantly higher number of
outliers than the autoencoder, it can be concluded that the autoencoder actually
narrows results of $K$-means with ponders indicating the most prominent outliers.

The obtained accuracy of the explainable surrogate model (the decision tree
trained according to the labelling induced by autoencoder based outliers)
computed on the training data 
is equal to 1 for both $L_{10}$ and $A_{10}$ datasets. This implies that
the surrogate models are able to explain without errors
why a particular instance is an outlier.
The explanation is given by the path from the root node of the 
decision tree to a leaf node determined by feature values of that instance.
The root node and low-depth nodes (nodes near to the root
node) of the decision are the most discriminative features of detected
outliers. Table~\ref{T3} shows the top ten most discriminative features
for $L_{10}$ and $A_{10}$ datasets together with their 
Gini importance scores.

\begin{table}[ht]
\centering
\caption{The top ten most discriminative features of outliers identified by the autoencoder.
$G$ denotes the Gini importance of a feature.}
\begin{tabular}{llll}
\noalign{\smallskip}\hline \noalign{\smallskip}
\multicolumn{2}{l}{$L_{10}$ dataset} & \multicolumn{2}{l}{$A_{10}$ dataset} \\
\noalign{\smallskip}\hline \noalign{\smallskip}
feature  & $G$ & feature & $G$ \\
\noalign{\smallskip}\hline \noalign{\smallskip}
{\it average salary in 16m4 }& 0.155 & {\it median salary in 16m8} & 0.369  \\
{\it fraction b26r in 16m12} & 0.121 & {\it average salary in 16m9} & 0.079 \\
{\it fraction b26r in 16m3} & 0.080 & {\it fraction b26r in 17m1} & 0.068 \\
{\it total employees in 16m7} & 0.078 & {\it stdev salary in 17m2} & 0.048 \\
{\it total persons in 16m3} & 0.069 & {\it fraction b26r in 17m3} & 0.047 \\
{\it stdev salary in 16m6} & 0.045 & {\it total persons in 16m12} & 0.044 \\
{\it total employees in 17m1} & 0.043 & {\it fraction b26r in 16m5} & 0.031 \\
{\it fraction b26r in 17m2} & 0.040 & {\it average age in 16m11} & 0.026 \\
{\it fraction b26r in 16m5} & 0.036 & {\it stdev salary in 16m4} & 0.021 \\
{\it org type 14} & 0.029 & {\it stdev salary in 17m3} & 0.020 \\
\noalign{\smallskip}\hline \noalign{\smallskip}
\end{tabular}
\label{T3}
\end{table}

The most discriminative feature for outliers in $L_{10}$ dataset
is the average salary of employees in April of 2016, {\it
average salary in 16m4}. The median salary in August of 2016,   {\it median salary in 16m8},
is the most discriminative features for outliers in $A_{10}$ dataset.
It can be observed that both lists contain a large number of
{\it b26r } related features introduced as domain knowledge features. Those features represent the fraction of
employees in the bin with the highest salary in Serbia. Also,
there are discriminative features related to the total number of 
persons associated to a business entities (regular employees plus
honorary and associate workers that do not have a permanent
working contract, {\it total employees in 16m7}, {\it total persons in 16m3}, {\it total employees in 17m1} ). Besides the mean and median, the standard deviation of 
salaries is also an important feature indicating outliers 
(in both datasets). The last feature in the list of top 10 features
discriminating $L_{10}$ outliers ({\it org type 14}) is a binary feature
indicating whether a business entity is a limited liability company.
The presence of this feature in the list implies that 
the probability that a randomly selected LLC is an outlier is higher 
than the probability that a randomly selected non-LLC is an outlier
for LLCs with less than 10 employees. 

Thus, the list of the most discriminative features reflects a tax arbitrage behavior as expected by relying on domain knowledge, like in feature {\it fraction b26r in YmN}. In the same time, the list captures some other less straightforward features which are more likely the product of a blind ML approach (like the dummy for LLC and the feature {\it total employees in 17m1} for L10 dataset or the feature {\it average age in 16m11}  for A10 dataset). Thanks to the blind approach, we find some features more than once i.e. in few specific time varieties. We may expect to have a high degree of correlation between time varieties of the same variable due to trend component. However, our hybrid model has probably discerned some peaks / irregularities in monthly level of payments of income as higher risk in terms of tax evasion by assigning high discriminatory power to few specific monthly varieties of some features (Table~\ref{T3}).

\subsection{Autoencoder Structure}

In the next experiment, we investigated the impact of the autoencoder's
structure to the resulting set of outliers. In addition to the previously
examined autoencoder with three hidden layers, outliers were also detected 
with an autoencoder
having one hidden layer and an autoencoder with five hidden layers. 
The autoencoder with one hidden layer contained $d/2$ hidden neurons.
The structural configuration of the autoencoder with 5 hidden layers
was $3d/4$, $d/2$, $d/4$, $d/2$ and $3d/4$ hidden neurons, sequentially 
per layer (recall that $d$ is the number of features). The parameters
describing the training process were identical as in the previous 
experiment. 

\begin{figure}[ht]
\centering
\begin{subfigure}{.5\textwidth}
  \centering
  \includegraphics[width=1\linewidth]{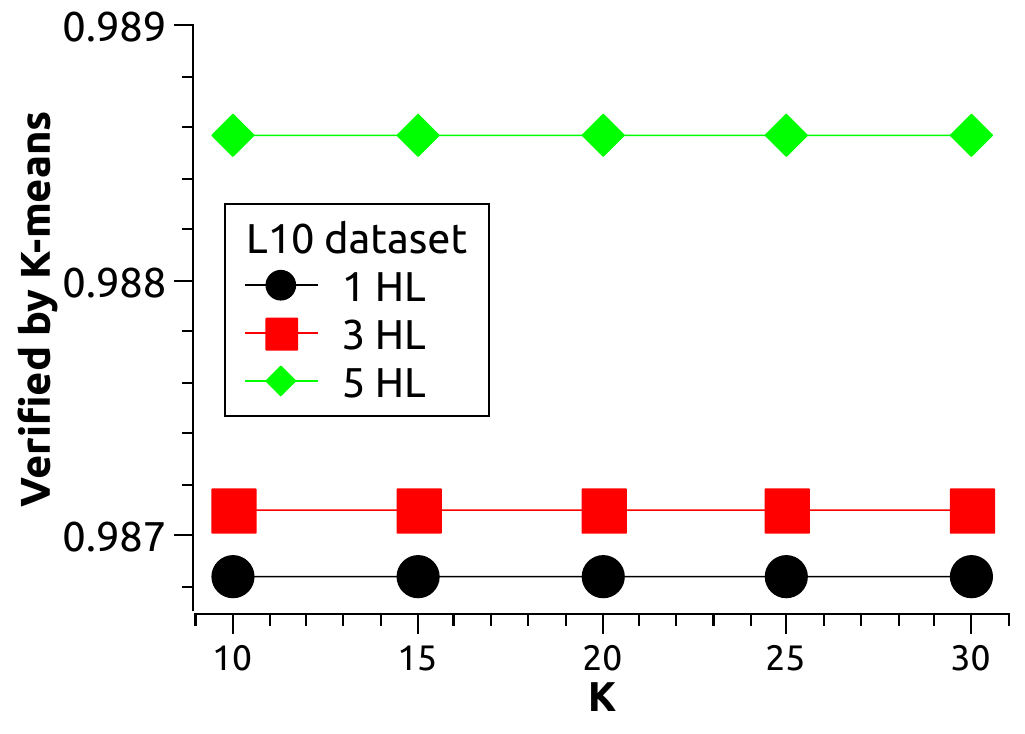}
  \caption{$L_{10}$ dataset}
  \label{fig2:sub1}
\end{subfigure}%
\begin{subfigure}{.5\textwidth}
  \centering
  \includegraphics[width=1\linewidth]{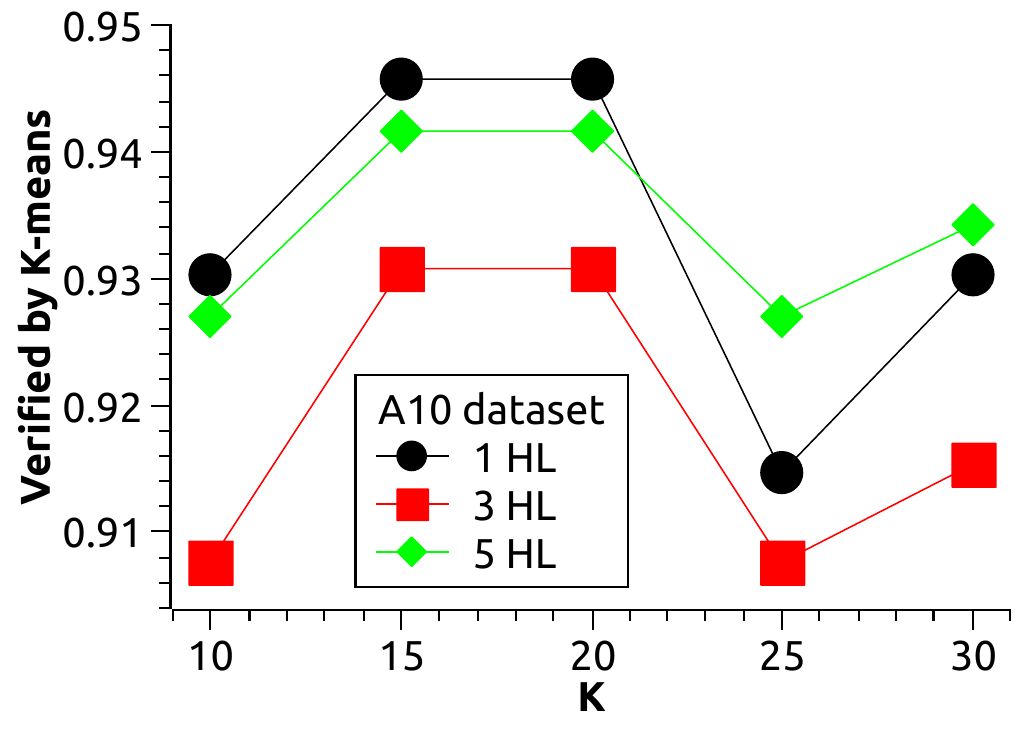}
  \caption{$A_{10}$ dataset}
  \label{fig2:sub2}
\end{subfigure}
\caption{The fraction of outliers verified by $K$-means with ponders for autoencoders with
1, 3 and 5 hidden layers and various K values.}
\label{fig2_aestruct}
\end{figure}

The fraction of outliers verified by $K$-means with ponders 
for all three examined autoencoders is shown in Figure~\ref{fig2_aestruct}.
For both datasets, the fraction of $K$-means verified
outliers is higher than 0.9 for various K values. More than 97\% of outliers 
detected by the autoencoders in $L_{10}$ are
also outliers indicated by $K$-means with ponders. The autoencoder with 5 hidden 
layers has the largest fraction of $K$-means verified outliers (98.83\% for all K values), then
the autoencoder with 3 hidden layers (98.71\% for all K values), followed by 
the autoencoder with 1 hidden layer (98.68\% for all K values).

In contrast to $L_{10}$ dataset, the fraction of $K$-means verified outliers in $A_{10}$ dataset 
depends on $K$. However, the differences in the number of $K$-means verified outliers for different 
$K$ values are less than 3.1\%. Small differences are also present for different autoencoders for 
a fixed $K$ (less than 3\%; for $K$ = 15 and $K$ = 20 less than 1.5\%). The autoencoder with one hidden layer has
the highest fraction of $K$-means verified outliers for $K \leq 20$. For larger $K$ values 
the autoencoder with 5 hidden layers exhibits the highest degree of agreement with
$K$-means with ponders.

\begin{figure}[ht]
\centering
\begin{subfigure}{.5\textwidth}
  \centering
  \includegraphics[width=0.9\linewidth]{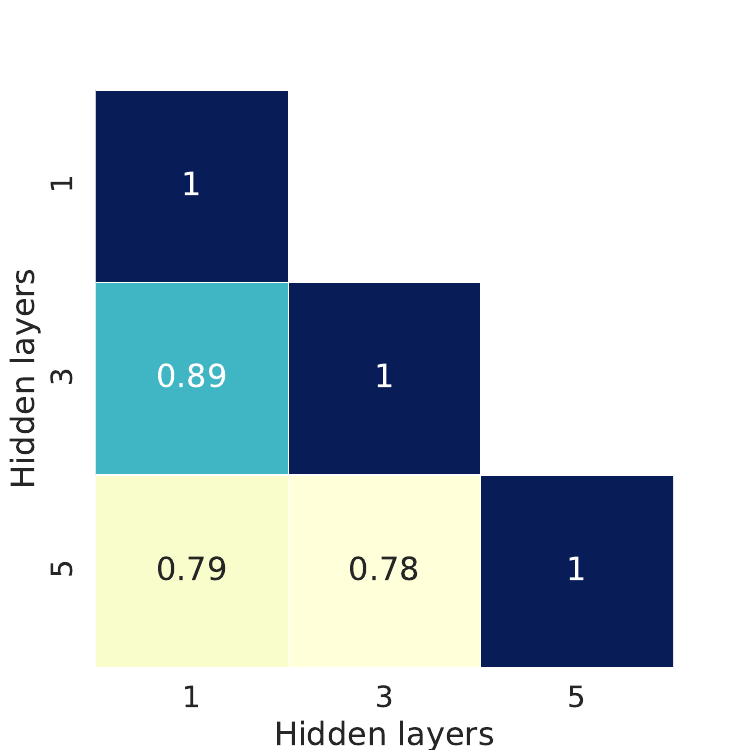}
  \caption{$L_{10}$ dataset}
  \label{fig3:sub1}
\end{subfigure}%
\begin{subfigure}{.5\textwidth}
  \centering
  \includegraphics[width=0.9\linewidth]{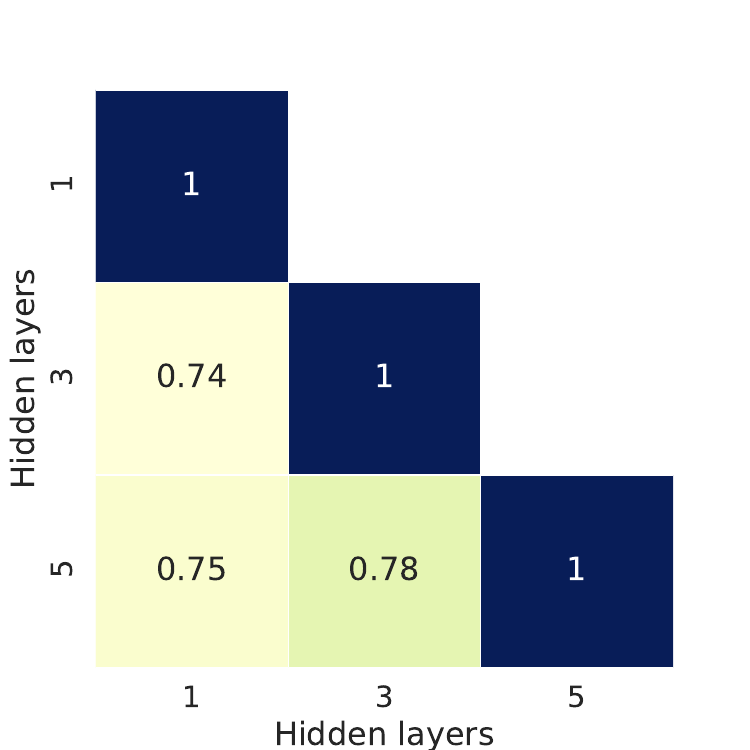}
  \caption{$A_{10}$ dataset}
  \label{fig3:sub2}
\end{subfigure}
\caption{The overlap between outlier sets for autoencoders with 1, 3 and 5 hidden layers measured
by the Jaccard coefficient.}
\label{fig3_aestruct_jaccrd}
\end{figure}

Figure~\ref{fig3_aestruct_jaccrd} shows the overlap between outlier sets measured by 
the Jaccard coefficient for each pair of examined autoencoders. It can be seen that
the lowest overlap is 0.75 implying a high degree of agreement in outliers detected
by structurally different autoencoders. Taking into account a high fraction of 
$K$-means verified outliers for structurally diverse autoencoders and a high degree 
of their mutual agreement, it can be concluded that the hybrid method 
is robust to variations in the structure of the autoencoder.

\subsection{Impact of Regularization}

In the last experiment we analyzed the impact of two regularization mechanisms aimed
to prevent autoencoder overfitting and make it more robust to unknown instances. 
Here we give the results obtained for the autoencoder with 3 hidden layers. 
The results for the autoencoders with 1 and 5 hidden layers are very similar to
those presented and discussed in this section and hence omitted here. 

\begin{figure}[htb!]
\centering
\includegraphics[width=0.7\linewidth]{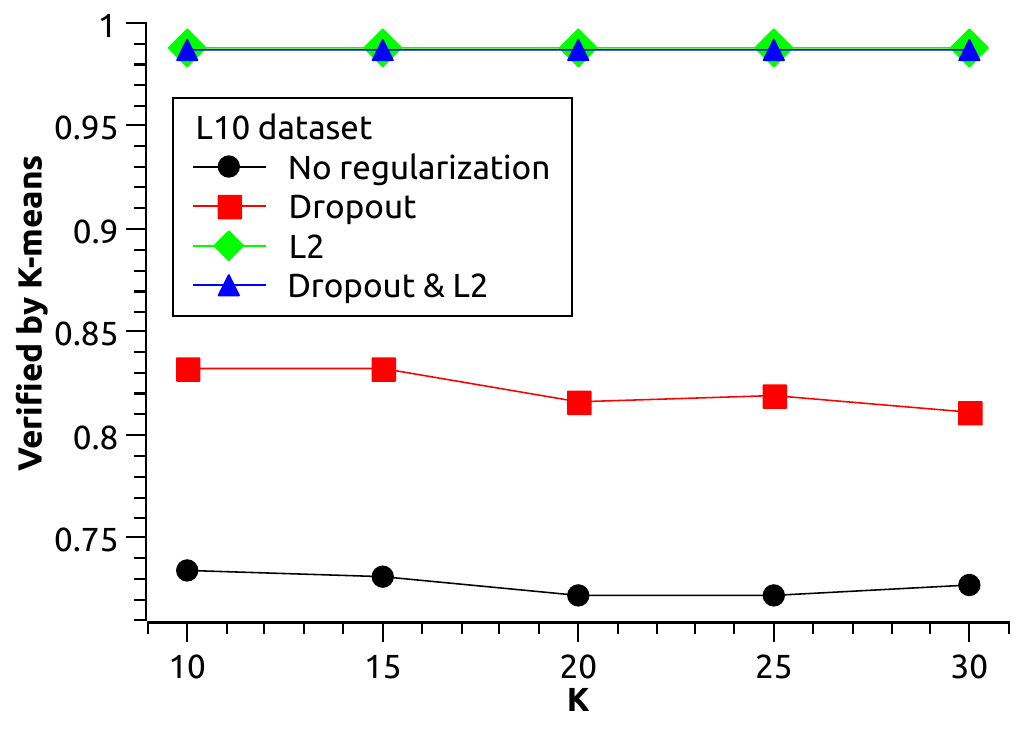}
  \caption{The impact of autoencoder regularization on outliers identified in $L_{10}$ dataset.}
\label{fig4_regimpact_l10}
\end{figure}

Figure~\ref{fig4_regimpact_l10} shows the fraction of $K$-means verified outliers in $L_{10}$ dataset
for four autoencoders: the autoencoder trained with both regularization mechanism (this 
autoencoder was examined in two previous experiments), the autoencoder trained using 
the $L_2$ activity regularization without dropout, the autoencoder trained using dropout
without the $L_2$ activity regularization and the autoencoder trained without 
regularization mechanisms. It can be observed that the fraction of $K$-means verified outliers
significantly drops down if the $L_2$ activity regularization is turned off. The autoencoder
trained with both regularization mechanisms behaves in the same way the autoencoder trained
only using $L_2$ activity regularization achieving more than 98\% $K$-means verified outliers.
On the other hand, the autoencoder trained only with dropout has less than 85\% of $K$-means
verified outliers, while the autoencoder without regularization identifies less than 75\% 
of $K$-means verified outliers.

\begin{figure}[htb!]
\centering
\includegraphics[width=0.7\linewidth]{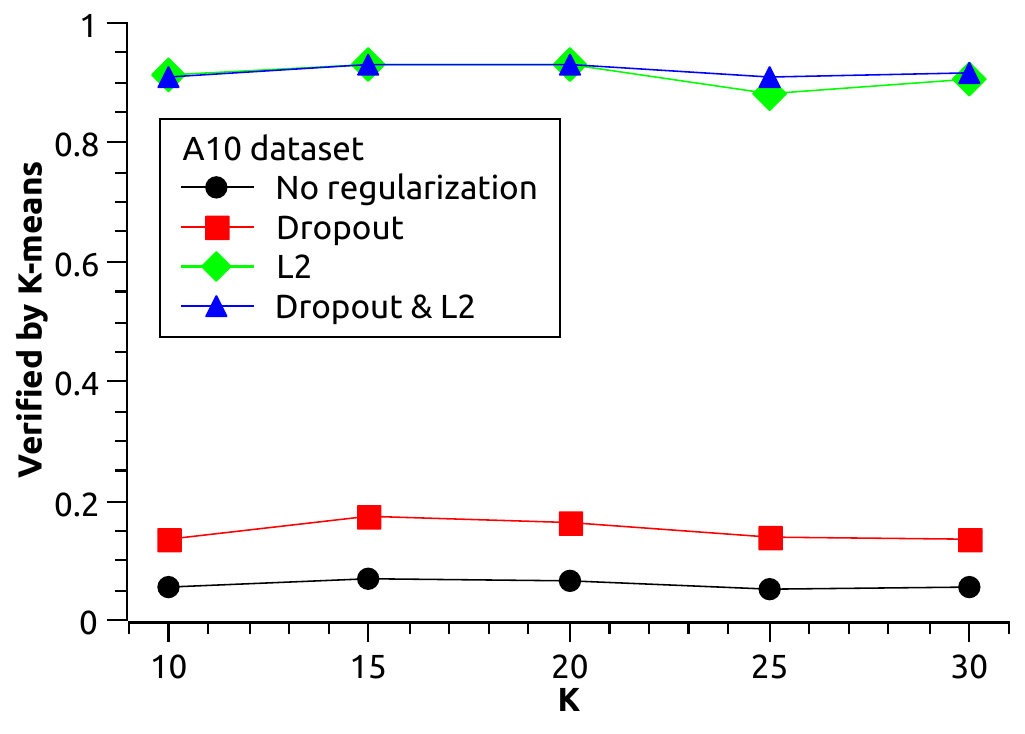}
  \caption{The impact of autoencoder regularization on outliers identified in $A_{10}$ dataset.}
\label{fig5_regimpact_l10}
\end{figure}

The importance of using the $L_2$ activity regularization when training the autoencoder 
is even more evident on $A_{10}$ dataset 
(Figure~\ref{fig5_regimpact_l10}). On this dataset, the fraction of $K$-means verified outliers 
drops from more than 90\% to less than 20\% when the $L_2$ activity regularization is deactivated.
The fraction of $K$-means verified outliers is less than 10\% when both regularization mechanisms
are excluded. Summarizing up the results obtained in this experiment, it can be concluded
that the  $L_2$ regularization mechanism leads to autoencoders with a  high 
degree of consistency with $K$-means with ponders significantly narrowing its result to outliers
indicated by two independent outlier detection approaches.

\subsection{Comparison to Alternative Outlier Detection Approaches}

To additionally demonstrate the effectiveness of HUNOD, we compare it with eight widely used outlier detection algorithms.
The following alternative outlier detection approaches provided by scikit-learn~\citep{scikit-learn} and PyOD~\citep{pyod} 
libraries are considered in our comparative analysis:
\begin{enumerate}
 \item SVM -- outlier detection by one-class support vector machines,
 \item LOF -- outlier identification based on local outlier factor,
 \item IF  -- isolation forest algorithm,
 \item EE  -- elliptic envelope algorithm,
 \item PCA -- outlier detection based on principal component analysis,
 \item HBOD -- histogram-based outlier detection,
 \item ABOD -- angle-based outlier detection, and
 \item KNN  -- outlier detection by the $K$-nearest neighbors algorithm.
\end{enumerate}
All considered alternatives identify outliers by computing an outlierness score to each data instance in a dataset.
Then, the instances are sorted by the outlierness score and the top $N$ results are returned as outliers, where $N$ is a threshold
specified by the user. Thus, we apply the alternative algorithms by setting $N$ to 
the number of outliers identified by the HUNOD autoencoder. Then, we check identified outliers against $K$-means with 
ponders for different values of $K$. Specifically for KNN and ABOD, outliers are identified with the same value of parameter $K$ that 
is used by $K$-means with ponders. The hyperparameters of other algorithms are set to their default values (as specified by scikit-learn
or PyOD).

Having in mind that $K$-means with ponders has a high recall, the fraction of outliers identified by an alternative method $A$ that are also 
outliers indicated by $K$-means with ponders can be considered as a measure of relative precision of $A$ (relative to $K$-means with ponders). 
If $A$ has a low value of the relative precision with respect to a model having high recall then it can be definitely stated that 
$A$ exhibits a poor outlier detection performance.

Table~\ref{TcompL10} shows the fraction of outliers identified by the alternative methods on $L_{10}$ that are also indicated 
as outliers by $K$-means with ponders for different $K$ values (the same as $K$ values used in Section~\ref{autoencoder_outliers}). 
It can be seen that LOF, PCA and ABOD have a very low relative precision on $L_{10}$. SVM, EE, HBOD and KNN reach a moderate level
of relative precision that is significantly lower than the relative precision of the HUNOD autoencoder (the precision of the 
HUNOD autoencoder on $L_{10}$ is equal to 0.987, please see Table~\ref{T2}). The highest relative precision on $L_{10}$ is 
achieved by IF, ranging between 0.981 (slightly lower than the HUNOD autoencoder) and 1.0 (slightly higher than the HUNOD autoencoder), 
i.e. the HUNOD autoencoder and IF identify nearly the same set of outliers.

\begin{table}[ht]
\centering
\caption{Outliers identified by alternative outlier detection approaches on $L_{10}$ checked
against $K$-means with ponders for different $K$ values.}
\begin{tabular}{lllllllll}
\noalign{\smallskip}\hline \noalign{\smallskip}
$K$ & SVM & LOF & IF & EE & PCA & HBOD & KNN & ABOD\\
\noalign{\smallskip}\hline \noalign{\smallskip}
10 & 0.549 & 0.013 & 0.981 & 0.690 & 0.116 & 0.439 & 0.794 & 0.006 \\
15 & 0.542 & 0.045 & 0.994 & 0.695 & 0.116 & 0.413 & 0.794 & 0.032 \\
20 & 0.542 & 0.006 & 1.000 & 0.712 & 0.116 & 0.439 & 0.800 & 0.006 \\
25 & 0.542 & 0.019 & 0.994 & 0.800 & 0.123 & 0.484 & 0.826 & 0.019 \\
30 & 0.490 & 0.013 & 0.987 & 0.779 & 0.116 & 0.445 & 0.545 & 0.019 \\
\noalign{\smallskip}\hline \noalign{\smallskip}
\end{tabular}
\label{TcompL10}
\end{table}

The relative precision of the alternative methods on $A_{10}$ dataset is given in Table~\ref{TcompA10}. Similarly as for
$A_{10}$, LOF, PCA and ABOD have a very low values of the relative precision. EE, HBOD and KNN also show poor outlier 
detection performance on $A_{10}$ with significantly lower values of the relative precision than on $L_{10}$. SVM and IF 
exhibit a moderate degree of consistency with $K$-means with ponders with the relative precision between 0.382 and 0.788.
This level of relative precision is significantly lower than the relative precision of the HUNOD autoencoder which is 
in the range [0.908, 0.931]. Thus, it can be concluded that the HUNOD autoencoder gives more precise results than 
the examined alternative outlier detection methods.

\begin{table}[ht]
\centering
\caption{Outliers identified by alternative outlier detection approaches on $A_{10}$ checked
against $K$-means with ponders for different $K$ values.}
\begin{tabular}{lllllllll}
\noalign{\smallskip}\hline \noalign{\smallskip}
$K$ & SVM & LOF & IF & EE & PCA & HBOD & KNN & ABOD\\
\noalign{\smallskip}\hline \noalign{\smallskip}
10 & 0.412 & 0.000 & 0.788 & 0.168 & 0.102 & 0.255 & 0.219 & 0.095 \\
15 & 0.434 & 0.000 & 0.796 & 0.139 & 0.102 & 0.241 & 0.190 & 0.162 \\
20 & 0.404 & 0.015 & 0.511 & 0.109 & 0.102 & 0.350 & 0.204 & 0.184 \\
25 & 0.382 & 0.015 & 0.507 & 0.080 & 0.095 & 0.299 & 0.146 & 0.190 \\
30 & 0.382 & 0.022 & 0.650 & 0.058 & 0.109 & 0.299 & 0.153 & 0.175 \\
\noalign{\smallskip}\hline \noalign{\smallskip}
\end{tabular}
\label{TcompA10}
\end{table}

\section{Conclusions and Future Work}
\label{conclusions}

In this paper, we proposed a novel approach for tax fraud risk management based on a hybrid machine learning-based approach that possesses several favorable features. First, the proposed method allows to incorporate the relevant domain knowledge into the model in a user friendly way, through setting of ponders (weights) of the different data features from a pre-defined set of a few candidate ponders. Second, the method offers an explainable decision tree model based on which domain experts can verify (manually or through another way) the anomalous entities that the ML model outputs. Hence, the proposed method allows for domain expert validation of the ML-declared anomalies. 
Finally, the method demostrates strong robustness in terms of anomaly validation due to its hybrid nature. Namely, anomalies obtained by two idependent anomaly detection methods ($K$-means and autoencoder) are cross-checked in order to devise the final set of internally-validated anomalies.

The experimental evaluation of the method shows that there is a very good embedding of outliers detected by the HUNOD autoencoder into much larger set of outliers detected by $K$-means with ponders. 
Given that these two outlier detection approaches are based on radically different machine learning designs and that the set of outliers detected by the autoencoder is significantly smaller and yet contained in the set of outlier detected by $K$-means to a very large degree, we conclude that the autoencoder actually refines the results of $K$-means indicating the most prominent outliers. 
Our experimental analysis also showed that the HUNOD autoencoder exhibits a higher degree of consistency with $K$-means than eight other alternative outlier detection algorithms. Since $K$-means exhibits a high recall, this result also suggests that the HUNOD autoencoder is a more precise outlier 
detection model than the examined alternatives.

The results of cross-checking and the results obtained by the decision tree further strengthen our initial idea that the integration of domain knowledge improves the ML approach significantly and offers the level of explainability not typical for ML methods. The features pondering appears to influence the results of $K$-means significantly while the most discriminative features are both expert-designed as the features aimed to capture the possible tax arbitrage as well as features without particular economic interpretation. Furthermore, the tests performed with different autoencoder structure and regularization parameters indicate that the HUNOD method is rather robust and that the $L_2$ regularization plays an important role in the method.

The presented method designed for tax fraud detection related to the personal income is contributing to the literature on tax fraud detection using big data methods where the prevailing literature is dealing with models for VAT evasion detection.
The further research could consist in additional development of the proposed hybrid model while the underlying dataset used for definition of features could be improved to encompass the relevant data related to the business entities (TINs) and individual persons from existing administrative and statistical data registries such as business financial statements, individual occupation, education level and length of working service. The dataset could be also complemented with VAT and profit tax for each business entity.

\begin{acknowledgements}
The authors would like to thank to Olivera Radi\v{s}a-Pavlovi\'{c} for her help
during the preparation of experimental data
and Nata\v{s}a Krklec-Jerinki\'{c} for proof-reading an initial version of this
article and giving us valuable comments and suggestions how to improve it.
The authors are also grateful to anonymous reviewers for their constructive and helpful comments and suggestions.

This research is supported by Ministry of Education, Science and Technological Development 
and Tax Administration of Republic of Serbia. The authors acknowledge financial support of 
the Ministry of Education, Science and Technological Development of 
the Republic of Serbia (Grant No. 451-03-9/2021-14/ 200125)
\end{acknowledgements}

%
%

\bibliographystyle{spbasic}      
\bibliography{hybridPU_refs}   

%
%

\end{document}